\documentclass[letterpaper, 10 pt, journal, twoside]{IEEEtran}  

\newcommand{\ve}[1]{\mbox{\boldmath $ #1 $}}
\newcommand{\dve}[1]{\dot{\mbox{\boldmath $ #1 $}}}
\newcommand{\ddve}[1]{\ddot{\mbox{\boldmath $ #1 $}}}

\usepackage[dvipdfmx]{graphicx}
\usepackage{bm}
\usepackage{amsmath}
\usepackage{amsfonts}
\usepackage{amssymb}
\usepackage{ascmac}
\usepackage{prettyref}
\usepackage{comment}
\usepackage{multirow}
\usepackage{makeidx}
\usepackage{fancyhdr}
\usepackage[caption=false,font=footnotesize]{subfig}
\usepackage{float}
\usepackage{cite}
\usepackage{color}
\usepackage{algorithm,algorithmic}
\usepackage{url}

\newrefformat{sec}{\ref{#1}“\nameref{#1}”}
\newrefformat{sub}{\ref{#1}“\nameref{#1}”}
\newrefformat{par}{\ref{#1}“\nameref{#1}”}
\newrefformat{tab}{\ref{#1}“\nameref{#1}”}
\newrefformat{fig}{Fig. \ref{#1}}

\IEEEoverridecommandlockouts                              

\title{\LARGE \bf
Stability analysis of admittance control \\ using asymmetric stiffness matrix
}

\author{Toshiaki Tsuji$^{1}$ and Yasuhiro Kato$^{1}$
\thanks{$^{1}$Toshiaki Tsuji and Yasuhiro Kato are with the 
Graduate School of Science and Engineering, Saitama University, 255 Shimo-okubo, Saitama, 338-8570 
{\tt\small E-mail: tsuji@ees.saitama-u.ac.jp}}%
}
\begin{document}

\maketitle
\thispagestyle{empty}
\pagestyle{empty}

\begin{abstract}
In contact-rich tasks, setting the stiffness of the control system is a critical 
factor in its performance. Although the setting range can be extended by making 
the stiffness matrix 
asymmetric, its stability has not been proven. This study focuses on the stability 
of compliance control in a robot arm that deals with an asymmetric stiffness matrix. 
It discusses the convergence stability of the admittance control. The paper explains 
how to derive an asymmetric stiffness matrix and how to incorporate 
it into the admittance model. The authors also present simulation and experimental 
results that demonstrate the effectiveness of their proposed method.
\end{abstract}

\section{Introduction}
In-demand automation tasks often involve contact-rich tasks with varying mechanical 
constraints between the robot and the environment, making control program 
implementation complex. Studies on contact-rich tasks using the peg-in-hole benchmark have expanded automation technology, particularly for precision assembly. 
Compliance mechanisms, such as active and passive compliance, help robots to perform assembly tasks despite positional errors. Impedance and admittance control are common 
methods for active compliance~\cite{hogan,newman}. 
Contact-model-based~\cite{tang, unten} and contact-model-free~\cite{IL1, RL1} approaches are used for compliance control, with the latter often using machine learning methods such as learning from demonstration and learning from the environment with reinforcement learning. 

Variable impedance control (VIC), a control strategy that adjusts a robot's mechanical impedance for safe, versatile, and efficient interaction, has been actively studied~\cite{vic_first}. VIC has shown promising results in various applications, such as manipulation, locomotion, haptic feedback, and rehabilitation. However, it faces several challenges, such as the design of robust and efficient actuators and sensors, as well as the development of effective control strategies for complex and dynamic tasks~\cite{vic_review}. 

Methods that use passive compliance obtain the desired stiffness of the hand 
by using a device that combines passive mechanical elements such as springs. 
The remote center compliance (RCC) device is one of the most popular examples, 
which utilizes the inter-axis interference to induce motion against external 
force~\cite{RCC}. Instead of VIC, variable RCC (VRCC), which can set stiffness 
in a variable manner, has been studied to deal with complicated tasks~\cite{VRCC1, VRCC2}. 

RCC has the advantage of not being limited by control bandwidth because it is 
composed of mechanical elements. On the other hand, when compliance is reproduced by control, 
even ranges that cannot be reproduced mechanically can be artificially 
reproduced with a single parameter setting. In other words, the compliance 
control has a wider range over which motion can be induced. However, how to 
set the parameters under an assurance of stability is an important issue. 

A passivity-based approach is the most popular candidate to ensure the stability 
of VIC~\cite{ferraguti2013}. It has also been extended to the time-varying 
admittance controller~\cite{ferraguti2015}. Kronander and Billard proposed 
stability conditions for the VIC scheme that works for different levels of 
stiffness and damping, using a modified Lyapunov function~\cite{human_stiff3}. 
Later, Sun {\it et al}. expanded on this idea and proposed new constraints 
that ensure both exponential stability and boundedness of the robot's position, 
velocity, and acceleration~\cite{sun}. 
Spyrakos-Papastavridis {\it et al}. proposed a Passivity-Preservation Control 
(PPC) that enables the implementation of stable VIC and provided joint and 
Cartesian space versions of the PPC controller permit the intuitive definition 
of interaction tasks~\cite{ppc}. 
In the study of parameter design for variable admittance control, Kim and Yang 
performed a stability analysis based on the root locus~\cite{admittance1}. 
The root locus has also been used in the parameter design for drill 
admittance control~\cite{admittance2}. 

While many stability analyses of compliance control have been conducted, 
only passivity-based analysis has been pursued in multi-degree-of-freedom systems because the problem becomes complex due to the consideration of interference in other axes~\cite{multi-dof}. 
Previous studies on compliance control, such as \cite{oikawa_RAL} and 
\cite{kozlovsky}, have indicated that introducing an asymmetric stiffness matrix 
can increase the robot's capability and extend the range of motion induction. 
However, the main challenge is that the asymmetric part fails to satisfy 
passivity, making it impossible to ensure stability. In this study, we derived 
parameters for the stability limit of the asymmetric stiffness matrix based on 
the eigenvalues of the system matrix. We then demonstrate that stability can 
be guaranteed by setting the parameters using the root locus method.

The contributions of this paper can be summarized as follows. 
\begin{enumerate}
\item
Spiral oscillations can be excited when the stiffness matrix is asymmetric. 
We have clarified the condition for such excitation. We also showed that the condition for the absence of spiral oscillation is equivalent to the condition that the eigenvalues of the stiffness matrix are real numbers. 
\item 
We have demonstrated that it is difficult to satisfy the passivity 
when the above conditions are not satisfied and therefore identified 
the conditions under which vibration converges using the root locus method.
\end{enumerate}
\section{Basic Principle}
\begin{figure}[tb]
    \centering
        \includegraphics[width=8.6cm]{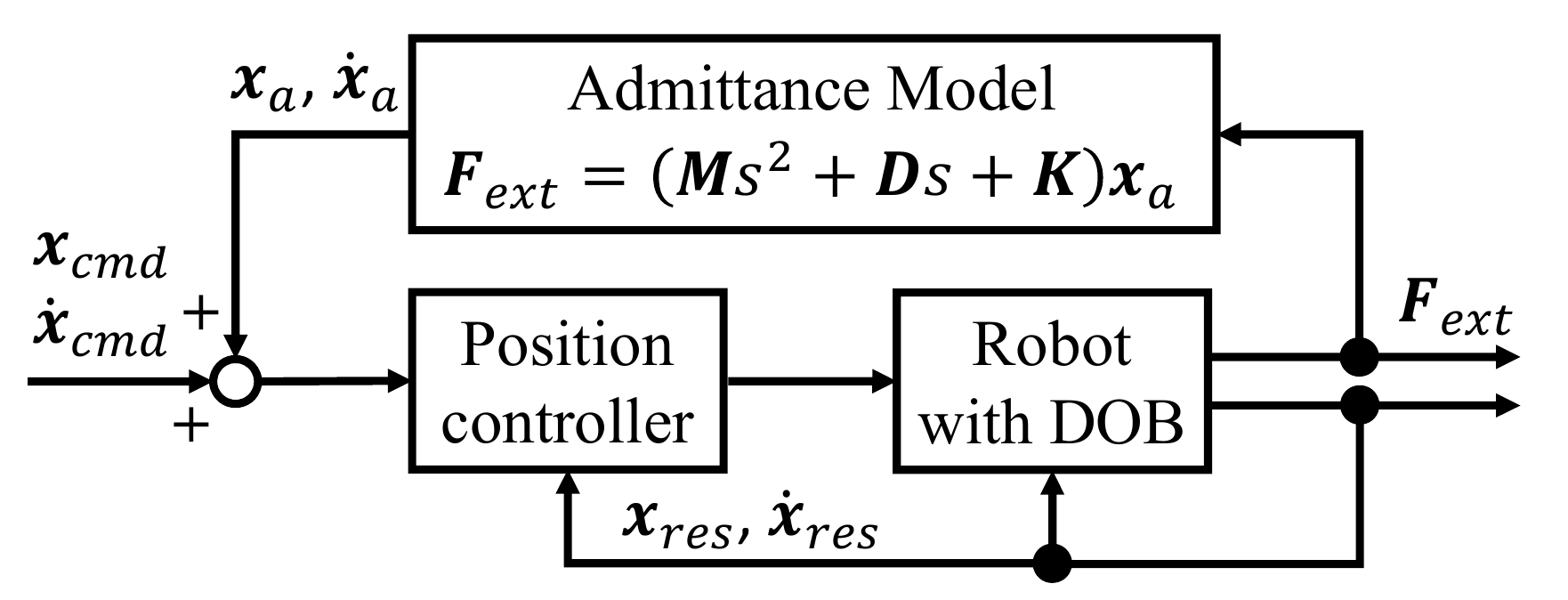}
        \caption{Admittance Control}
        \label{fig:block_diagram}
\end{figure}

This study discusses the stability of compliance control in a robot arm that 
deals with an asymmetric stiffness matrix. It is limited to admittance control 
and discusses the convergence stability of a planar admittance model, assuming the 
inner position control loop is stable. 
The feedback control system 
is shown in Fig.~\ref{fig:block_diagram}. It is an admittance control system with 
a minor feedback loop for position control. The type of position control is not specified, 
but a control system using a disturbance observer (DOB) is adopted in the experimental 
setup. The admittance model is based on the following equation: 
\begin{eqnarray}
  \ve{F}_{ext}=\ve{M}\ddve{x}+\ve{D}\dve{x}+\ve{K}\ve{x}
  \label{motion_equation}
\end{eqnarray}
This admittance model takes force $\ve{F}_{ext}$ as input and outputs the position 
$\ve{x}$ and velocity $\dve{x}$ obtained through numerical integration based on 
the acceleration $\ddve{x}$ obtained from equation (\ref{motion_equation}). 
While symmetric matrices are typically used for the stiffness matrix $\ve{K}$ and damping matrix $\ve{D}$ of the admittance model, this study considers the use of an asymmetric matrix for the stiffness matrix $\ve{K}$. For simplicity, the damping matrix $\ve{D}$ is assumed to be a diagonal matrix.

When an external force occurs, the direction in which the robot end effector is induced, and the convergence point are determined only by the stiffness matrix $\ve{K}$ of the admittance model. Therefore, it is desirable to expand the selection range of the stiffness matrix $\ve{K}$ when designing the relationship between external forces and robot operation. On the other hand, since the damping matrix $\ve{D}$ does not affect the convergence point of the end effector, the priority of expanding the selection range is not high. 
Additionally, applying viscous force only in the direction in which velocity is 
generated increases the dissipation of energy and is reasonable for enhancing stability. 
Therefore, it is necessary to consider how to give the diagonal elements 
of the damping matrix $\ve{D}$ when the non-diagonal elements of the stiffness matrix 
$\ve{K}$ are given arbitrarily. 

\begin{figure}[tb]
    \centering
        \includegraphics[width=8.6cm]{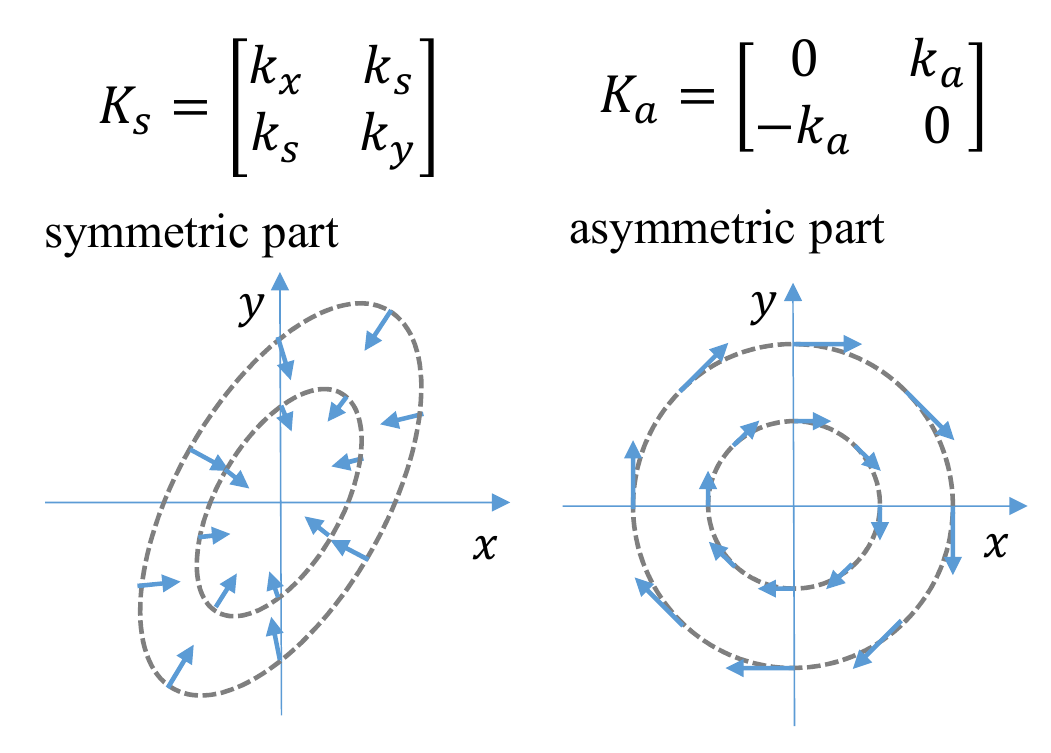}
        \caption{Force field generated by asymmetric stiffness matrix}
        \label{fig:force_field}
\end{figure}

When the stiffness matrix $\ve{K}$ is asymmetric, it is known that the effect of 
the stiffness matrix can be expressed as a superposition of the symmetric part 
$\ve{K}_s$, which creates a force field that converges at the equilibrium point~\cite{hogan}, 
and the asymmetric part $\ve{K}_a$, which creates a force that rotates around 
the equilibrium point. An example of this is shown in Fig.~\ref{fig:force_field}. 
If the stiffness control is reproduced by a virtual spring, the symmetric part can be treated as generating an elliptical virtual potential energy field generated by strain. Similarly, virtual kinetic energy can be derived from the mass and velocity of the admittance model. Therefore, in the symmetric matrix, a control system design based on passivity can be achieved by deriving an energy 
function. However, when this is extended to an asymmetric matrix, the curl 
field generated by the asymmetric part is not a conservative force, and it may excite 
oscillations. Oscillations excited by the curl field are referred to as spiral oscillations. 
Since energy may increase due to spiral oscillation, 
passivity may not be satisfied. 

This is demonstrated by the energy function. It is obtained as the integral 
of the product of velocity and force, which is power. 
\begin{eqnarray}
    V\!\!&=&\int \dve{x}^T \ve{F}_{ext}dt \label{eq:integralpower}
\end{eqnarray}
By substituting $\ve{K}=\ve{K}_s+\ve{K}_a$ and (\ref{motion_equation}) 
into (\ref{eq:integralpower}), 
\begin{eqnarray}
   V\!\!&=&\int \dve{x}^T \ve{M}\ddve{x} + \dve{x}^T \ve{D}\dve{x} + \dve{x}^T \ve{K}_s\ve{x} + \dve{x}^T \ve{K}_a\ve{x}\\
  \!\!&=&\!\!\!\! \frac{1}{2}\dve{x}^T\!\!\ve{M}\dve{x} + \frac{1}{2}\ve{x}^T\!\! \ve{K}_s\ve{x} + 
        \!\!\int\!\! \dve{x}^T\!\! \ve{D}\dve{x} dt + \!\!\int\!\! \dve{x}^T\!\! \ve{K}_a\ve{x} dt \label{eq:4terms}
\end{eqnarray}
Eq.~(\ref{eq:4terms}) shows that the virtual energy of the admittance model is defined as a function divided 
into four terms: kinetic energy, stiffness energy, energy loss due to friction, 
and energy imparted by the asymmetric part. 
The first and second terms represent kinetic and potential energy, respectively. 
The third term represents energy dissipated by viscosity, while the fourth term 
represents energy imposed by the curl field. Since the first and second terms are the energy of conservative forces, if the third term is smaller than the fourth term, the energy 
function does not increase. However, the above equation depends on velocity, 
and when the velocity decreases, the dissipative term becomes smaller. 
Therefore, if there are periods during vibration where the velocity becomes 
small, passivity may not hold. 

\section{Stability analysis using root locus}
The fundamental principle of admittance control described in the previous section shows 
that it is generally difficult to establish passivity when the stiffness matrix is 
made asymmetric. Therefore, the stability will be analyzed based on the concept of 
root locus below. First, the admittance parameters are defined in the x-y plane as follows:
\begin{equation}
  \ve{K} = \ve{K}_s + \ve{K}_a = \begin{bmatrix}
    k_x & k_s+k_a \\
    k_s-k_a & k_y 
  \end{bmatrix}, \ve{D} = \begin{bmatrix}
    d & 0 \\
    0 & d 
  \end{bmatrix}
\end{equation}
we define the state vector $\ve{x}$ as 
$[x_x, x_y, \dot{x}_x, \dot{x}_y]$, then its state equation can be obtained as follows:
\begin{eqnarray}
  \dve{x} &=& \ve{A}\ve{x} \nonumber \\
  \begin{bmatrix}
    \ddot{x}_x \\ \ddot{x}_y \\ \dot{x}_x \\ \dot{x}_y
  \end{bmatrix} &=& 
  \begin{bmatrix}
    -\frac{d}{m} & 0 & -\frac{k_x}{m} & \frac{-k_s-k_a}{m} \\
    0 & -\frac{d}{m} & \frac{-k_s+k_a}{m} & -\frac{k_y}{m} \\
    1 & 0 & 0 & 0 \\
    0 & 1 & 0 & 0
  \end{bmatrix}
  \begin{bmatrix}
    \dot{x}_x \\ \dot{x}_y \\ x_x \\ x_y
  \end{bmatrix}
\end{eqnarray}
Here, $k_x,k_y,k_s,k_a,d,m\in \ve{R}$. $d$ and $m$ are the damper and mass of the admittance 
model. $k_x$ and $k_y$ represent the diagonal elements of the stiffness matrix in the 
x and y directions, respectively, and $k_s$ and $k_a$ represent the symmetric and 
asymmetric components of the non-diagonal terms, respectively. Moreover, we define 
parameters as follows:
$d_m=\frac{d}{m},k_{xm}=\frac{k_x}{m}, k_{ym}=\frac{k_y}{m}, k_{sm}=\frac{k_s}{m}, k_{am}=\frac{k_a}{m}$, 
then we can write $\ve{A}$, the system matrix of the state equation, as follows:
\begin{eqnarray}
  \ve{A} &=& 
  \begin{bmatrix}
    -d_m & 0 & -k_{xm} & -k_{sm}-k_{am} \\
    0 & -d_m & -k_{sm}+k_{am} & -k_{ym} \\
    1 & 0 & 0 & 0 \\
    0 & 1 & 0 & 0
  \end{bmatrix} 
\end{eqnarray}

\begin{figure}[tb]
    \centering
    \includegraphics[width=8.6cm]{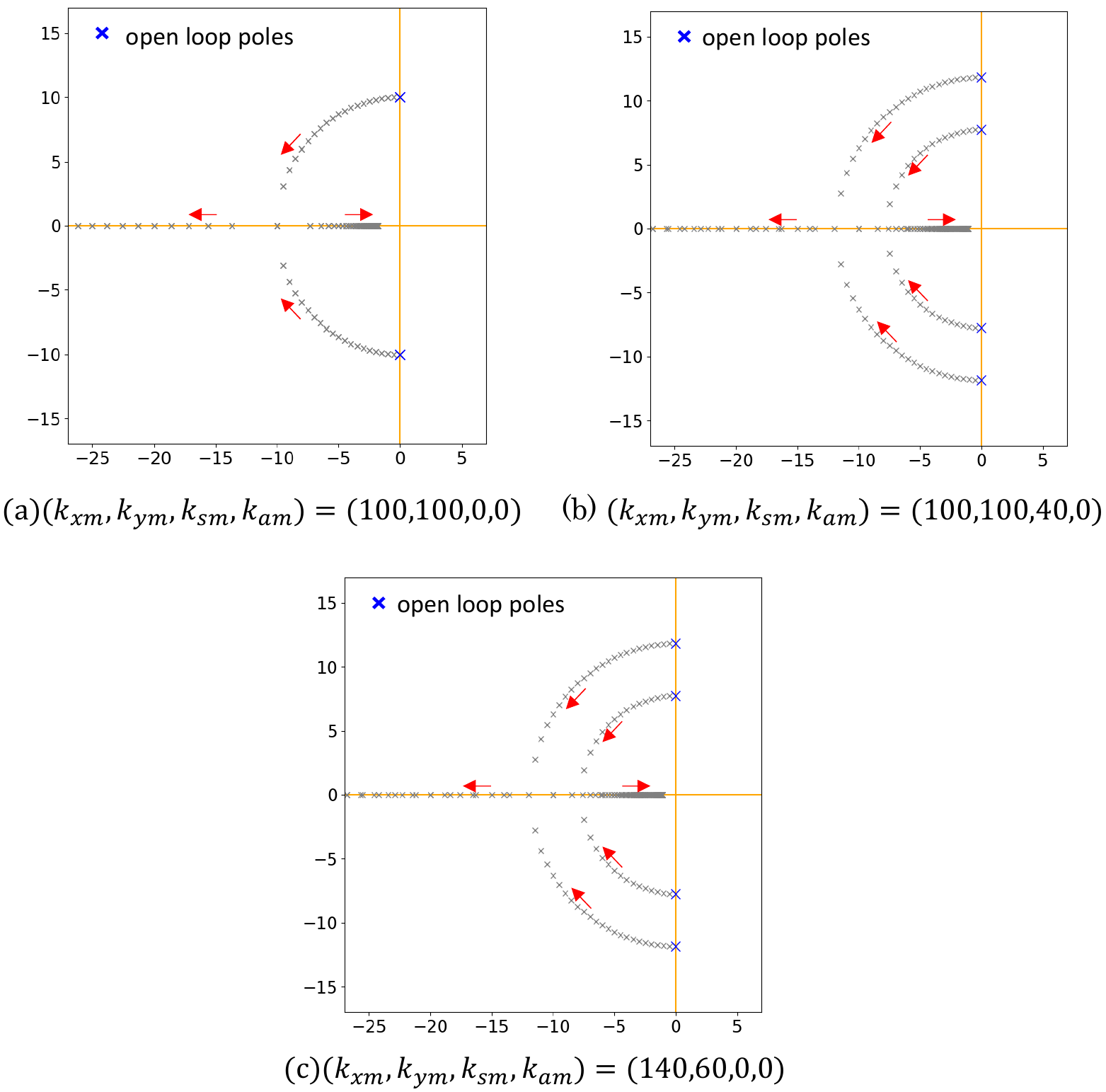}
    \caption{The root locus of a system with a symmetric stiffness matrix}
    \label{fig:rl1}
\end{figure}

The eigenvalues of the system matrix $\ve{A}$ are obtained as follows:
\begin{equation}
  \!\!\lambda\!=\!\!\frac{\!\!-d_m\!\!\pm \!\!\sqrt{\!d_m^2\!\!-\!\!2(k_{xm}\!\!+\!\!k_{ym}\!\!\pm\!\!
  \sqrt{\!\!-4k_{am}^2\!\!+\!\!4k_{sm}^2\!\!+\!\!(k_{xm}\!\!-\!\!k_{ym})^2\!})}}{2}
  \label{eq:lambdaa}
\end{equation}
The poles of the transfer function of this admittance model have the same values 
as the eigenvalues of the system matrix $\ve{A}$. Therefore, the root locus can 
be obtained by sweeping a parameter of matrix $\ve{A}$ from 0 to infinity. 
When the non-diagonal element $k_{am}$ of the stiffness matrix is zero, 
i.e., when the stiffness matrix is symmetric, the root locus for varying 
the damping ratio $d_m$ from 0 to infinity is shown in Fig.~\ref{fig:rl1}. 
For a symmetric matrix, the roots of the natural vibrations, which depend on the stiffness matrix, appear on the imaginary axis, and as the viscous damping ratio $d_m$ increases, the roots move in the negative real axis direction. 
At the same time, the imaginary part of the root decreases. 
The damping ratio $d_m$ where the imaginary part of the two conjugate 
eigenvalues shrinks to zero is called the critical damping ratio. 
The contents of the square root of (\ref{eq:lambdaa}) are then zero, 
hence the critical damping ratio is derived by the following equation.
\begin{eqnarray}
  d_m\!\!= \!\sqrt{2\left(k_{xm}\!+\!k_{ym}\!\!\pm\!\!
  \sqrt{\!-4k_{am}^2\!\!+\!4k_{sm}^2\!\!+\!(k_{xm}\!\!-\!k_{ym})^2}\right)}
\end{eqnarray}
When the value is equal to the critical damping ratio, the roots coincide on 
the real axis. 
When the viscous damping ratio 
$d_m$ increases beyond this value, the roots separate and move to the left 
and right along the real axis. 

Fig.~\ref{fig:rl1}(a) shows the result when $k_{xm}=k_{ym}=100$ and $k_{sm}=0$, 
and the stiffness matrix is diagonal. On the other hand, Fig.~\ref{fig:rl1}(b) 
shows the result when $k_{xm}=k_{ym}=100$ and $k_{sm}=40$. In other words, 
the stiffness matrix is not diagonal and its eigenvalues are 140 and 60. 
Fig.~\ref{fig:rl1}(c) shows the result for the case of a diagonal stiffness 
matrix with the same eigenvalues as in Fig.~\ref{fig:rl1}(b), with 
$k_{xm}= 140,k_{ym}=60, k_{sm}=0$. 
The initial position of the root locus changes depending on the eigenvalues of 
the stiffness matrix, as shown in the comparison of Figs.~\ref{fig:rl1}(a), (b), and (c). 

In Fig.~\ref{fig:rl1}(a), since the two eigenvalues coincide, the root locus 
starts from the points $\pm\sqrt{k_{xm}} j$ on the imaginary axis, while in 
Fig.~\ref{fig:rl1}(b), the root locus is split into two upper and lower 
branches due to the different eigenvalues. Figs.~\ref{fig:rl1}(b) and (c) 
also show that the root locus coincides when the eigenvalues are the same. 
Eq. (\ref{motion_equation}) deals with a 
generalized stiffness matrix in two dimensions, but to simplify the 
derivation of the equation, $k_{sm}$ is set to 0, and stability analysis 
is performed below. For the case where $k_{sm}$ is not 0, the same 
root locus can be obtained by deriving its eigenvalues and substituting 
them into the diagonal elements. 

\begin{figure}[tb]
    \centering
    \includegraphics[width=8.6cm]{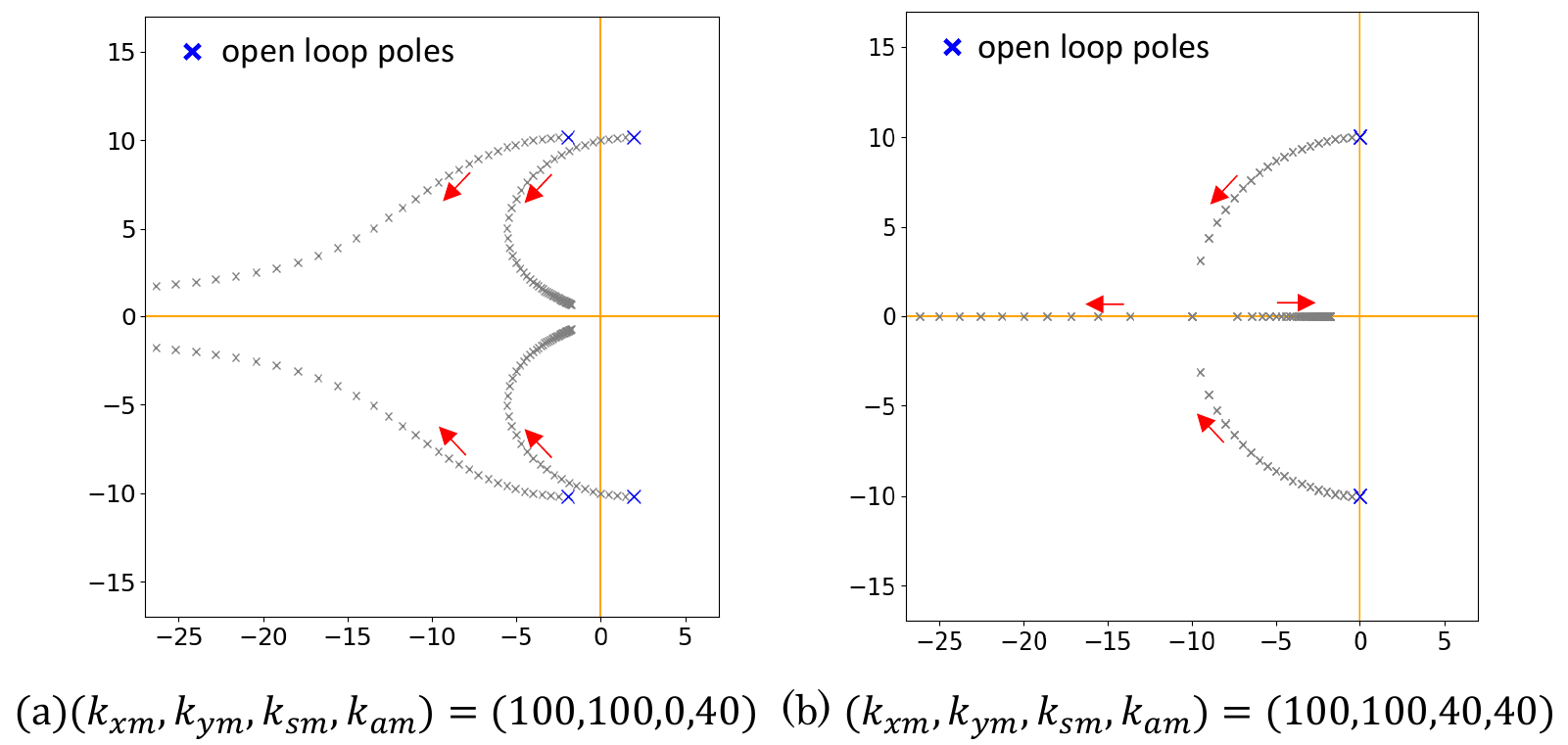}
    \caption{The root locus of a system with an asymmetric stiffness matrix}
    \label{fig:rl2}
\end{figure}

The root locus of the asymmetric matrix is shown in Fig.~\ref{fig:rl2}, 
where four roots appear symmetrically across the imaginary axis when $d_m$ is 0. 
As $d_m$ is gradually increased from 0, the four roots move negatively along 
the real axis while the imaginary axis component decreases. 
As $d_m$ is further increased, two of the four roots change their orientation 
toward the positive direction of the real axis, and the other two roots move 
in opposite directions, one positive and one negative. In this case, the imaginary 
axis component remains slightly. This means that in the case of an asymmetric 
matrix, oscillations are excited due to the presence of the curl field. 
However, there is a condition for the parameter that excites the oscillation. 
To ensure that all eigenvalues $\lambda$ are real, it is necessary for the 
contents of the square root in the second term of the numerator in 
(\ref{eq:lambdaa}) to be positive, 
that is,
\begin{equation}
  d_m^2\!-2\!\left(\!k_{x m}\!+\!k_{y m}\!\pm\!\!
\sqrt{\!-4 k_{a m}^2\!+4 k_{s m}^2\!+\!\left(k_{x m}\!-\!k_{y m}\right)^2}\right)\!\!>\!0.
  \label{eq:dm2}
\end{equation}
And it is necessary to have a real value of $d_m$ in (\ref{eq:dm2}). 
To ensure this, the square root term in (\ref{eq:dm2}) must be positive. 
The condition is expressed by the following inequality.
\begin{equation}
  -4k_{am}^2+4k_{sm}^2+(k_{xm}-k_{ym})^2>0 \label{eq:11}
\end{equation}
Consequently, condition for the asymmetric element $k_{am}$ is determined as follows:
\begin{equation}
  |k_{am}| < \sqrt{k_{sm}^2+1/4 (k_{xm}-k_{ym})^2}
  \label{eq:kam}
\end{equation}
If the above inequality is satisfied, the oscillation will no longer occur when 
$d_m$ is greater than the critical damping.
The eigenvalues of the stiffness matrix $\ve{K}$ are derived by the following equation:
\begin{equation}
  \lambda_K=\frac{1}{2}\left(k_{x m}+k_{y m} \pm \sqrt{-4 k_{a m}^2+4 k_{s m}^2+\left(k_{x m}-k_{y m}\right)^2}\right)
  \label{eq:lambdak}
\end{equation}
The condition for the stiffness matrix to obtain critical damping is equivalent to 
the matrix having real eigenvalues, as shown by the fact that the condition for 
(\ref{eq:lambdak}) to have real solutions is the same as that for (\ref{eq:kam}). 

Next, we will derive the conditions for the admittance parameters that ensure stability. 
If the viscous damping coefficient $d_m$ is increased to a certain value or more, 
all the roots will move toward the left half of the complex plane, making it possible 
to stabilize the control system. From (\ref{eq:lambdaa}), in the case of an asymmetric matrix,
\begin{eqnarray}
k_{xm}+k_{ym}\pm\sqrt{(-4k_{am}^2+(k_{xm}-k_{ym})^2}<0                        
\label{eq:kxmkym}
\end{eqnarray}
The above condition results in the occurrence of positive eigenvalues and unstable poles.
However, if $-4k_{am}^2+(k_{xm}-k_{ym})^2>0$, there are no solutions that satisfy 
(\ref{eq:kxmkym}) for $k_{xm}>0,k_{ym}>0$. Therefore, the condition for unstable poles to occur 
within the range of $-4k_{am}^2+(k_{xm}-k_{ym})^2\leq0$ needs to be investigated. 
To simplify the (\ref{eq:lambdaa}), $\alpha, \beta \in \ve{R}$ are defined as follows:
\begin{eqnarray}
\alpha=\!d_m^2\!-\!2(k_{xm}\!+\!k_{ym}),\beta=\!\!\sqrt{16k_{am}^2\!-\!4(k_{xm}\!-\!k_{ym})^2}. 
\label{eq:alphabeta}
\end{eqnarray}
Then, the eigenvalues are expressed as follows:
\begin{eqnarray}
\lambda=\frac{1}{2} (-d_m\pm\sqrt{\alpha\pm\beta j})                                           
\end{eqnarray}
Let the real part and imaginary part of $\sqrt{\alpha\pm\beta j}$ be $a$ and $b$, respectively, then
\begin{eqnarray}
(a+bj)(a+bj)=a^2-b^2+2abj=\alpha+\beta j                          
\end{eqnarray}
Consequently, $\alpha= a^2-b^2$, and $\beta=2ab$ are obtained. 
By the following equation: $\alpha+\sqrt{\alpha^2+\beta^2}=2a^2$, 
real part $a$ is expressed as below:
\begin{eqnarray}
a=\pm\sqrt{\frac{\alpha+\sqrt{\alpha^2+\beta^2}}{2}}                                          
\label{eq:a}
\end{eqnarray}
In order to avoid unstable poles, it is necessary for all eigenvalues to be negative, 
so we only consider the case where 
$a=\sqrt{\frac{\alpha+\sqrt{\alpha^2+\beta^2}}{2}}$. 
\begin{eqnarray}
-d_m+\sqrt{\frac{\alpha+\sqrt{\alpha^2+\beta^2}}{2}}<0 
\label{eq:dmpsqrt}
\end{eqnarray}
The condition that satisfies the (\ref{eq:dm2}) is derived below. First, consider the 
typical case where the equation is simplified, which is $k_{xm}=k_{ym}$. 
Substituting $\alpha=d_m^2-4k_{xm}$, $\beta=4k_{am}$ into (\ref{eq:dmpsqrt}) and 
the following equation is obtained:
\begin{eqnarray}
d_m>\sqrt{\frac{k_{am}^2}{k_{xm}}}
\label{eq:dmlimit}
\end{eqnarray}
If the value of $k_{am}$ is greater than $k_{xm}$, then the lower limit of $d_m$ also increases. 
These parameters are determined by the mass ratio, and to convert them into actual admittance parameters, we substitute $k_{xm}=k_x/m,k_{am}=k_a/m$, and $d_m=d/m$ into the above equation, yielding the following equation:
\begin{eqnarray}
d>\sqrt{\frac{k_a^2 m}{k_x}}
\label{eq:dlimit}
\end{eqnarray}
The next is to consider the case where $k_{ym}>k_{xm}>0$ to further generalize 
(\ref{eq:dmlimit}) and (\ref{eq:dlimit}). 
The condition is still given by (\ref{eq:alphabeta}), but now we substitute 
the more generalized (\ref{eq:dm2}). 
It is difficult to algebraically derive the condition for all eigenvalues to be negative. 
However, if (\ref{eq:dmlimit}) is satisfied, then all eigenvalues will be negative. 

The proof is shown below. If (\ref{eq:dmpsqrt}) is satisfied and without 
changing the values of the other parameters, let us increase the value of $k_{ym}$ from $k_{xm}$. 
If $a$ in (\ref{eq:a}) does not increase during this process, (\ref{eq:alphabeta}) will 
always be satisfied. When differentiating (\ref{eq:a}) with respect to $\alpha$ and $\beta$, 
we obtain the following expressions: 
\begin{eqnarray}
\frac{\partial a}{\partial\alpha}&=&\frac{1+\frac{a}{\sqrt{\alpha^2+\beta^2}}}
{2\sqrt{2}\sqrt{\alpha+\sqrt{\alpha^2+\beta^2}}}
\\
\frac{\partial a}{\partial\beta}&=&\frac{\beta}
{2\sqrt{2}\sqrt{\alpha^2+\beta^2}\sqrt{\alpha+\sqrt{\alpha^2+\beta^2}}}
\end{eqnarray}
when $\beta>0$ and $k_{ym}>k_{xm}$, both partial derivatives are positive. 
As (\ref{eq:alphabeta}) shows, increasing $k_{ym}$ always results in a decrease in both $\alpha$ and $\beta$, 
so the value of $a$ in (\ref{eq:a}) must always decrease. Therefore, if (\ref{eq:dmlimit}) is satisfied, 
then for all $k_{ym}>k_{xm}>0$, all eigenvalues are always negative. 
Replacing $k_{xm}$ and $k_{ym}$ in (\ref{eq:dmlimit}) as follows:
\begin{eqnarray}
d_m>\sqrt{\frac{k_{am}^2}{k_{ym}}} \label{eq:24}
\end{eqnarray}
By performing the same equation expansion as above, stability can also be 
demonstrated for the case where $k_{xm}>k_{ym}>0$. 

\section{Simulation study of convergence}
\subsection{Comparison of time response}
\begin{figure}[tb]
    \centering
        \includegraphics[width=8.6cm]{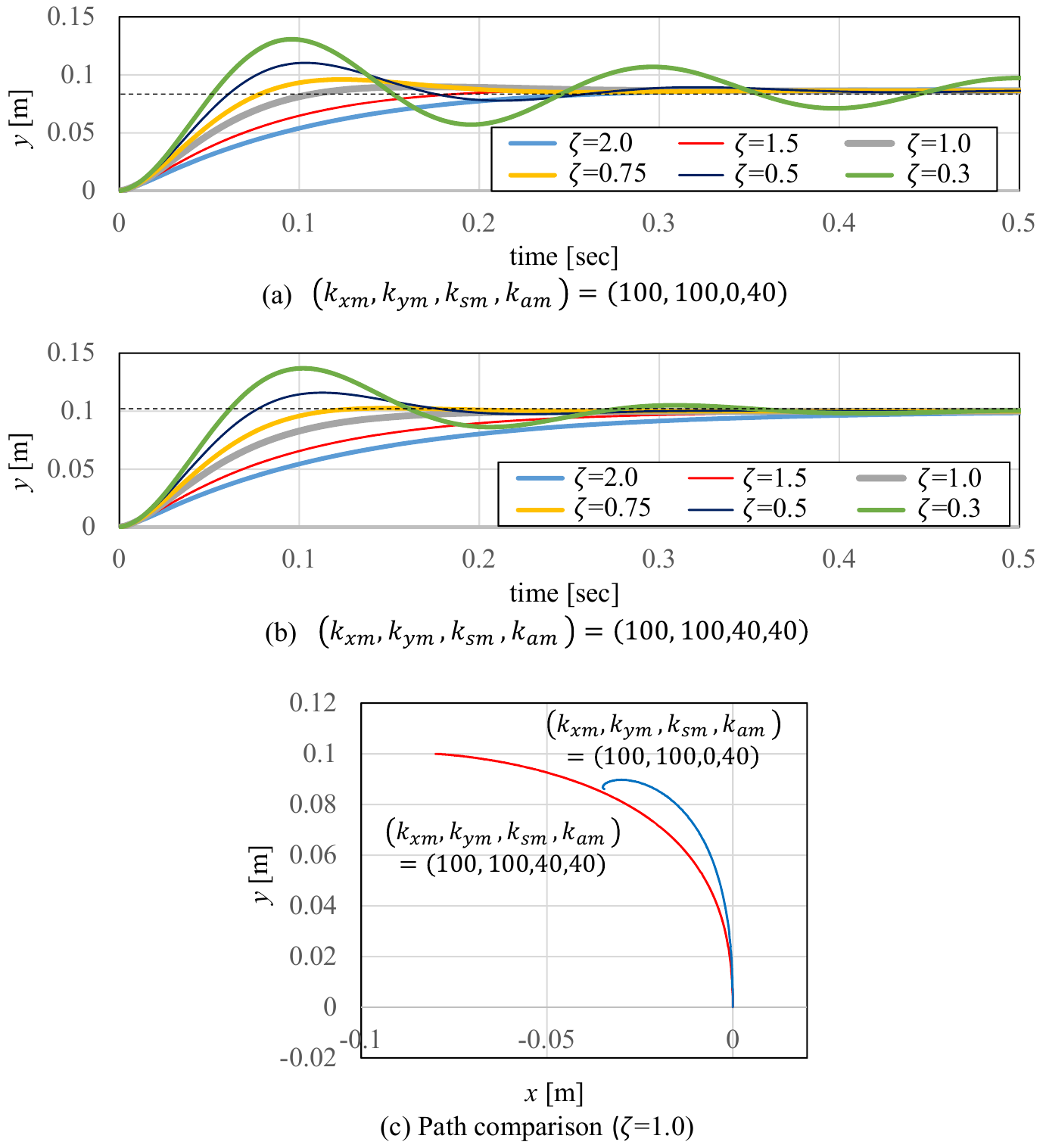}
        \caption{Time response of asymmetric matrix }
        \label{fig:time_responses}
\end{figure}

A simulation study on the convergence of the proposed method's root locus was 
conducted to examine its relationship with stability. 
An external force of $F_{ext}=(0.0,10.0)^T$ was step-inputted to the admittance 
model on a 2D plane at equilibrium. 
The damping coefficient $d_m$ was derived as follows and a comparative study 
was conducted while changing the value of the damping ratio $\zeta$. 
\begin{equation}
  d_m\!=\zeta \sqrt{2\!\left(\!k_{x m}\!+\!k_{y m} \!\pm \!\!\sqrt{\!\!-4 k_{a m}^2\!+\!4 k_{s m}^2\!+\!\left(k_{x m}\!\!-\!k_{y m}\right)^2}\right)}
\end{equation}
Fig.~\ref{fig:time_responses}(a) illustrates the simulation results conducted under the same conditions 
as those in Fig.~\ref{fig:rl2}(b). The convergence of the system is determined by the 
satisfaction of (\ref{eq:dlimit}). As the damping ratio $d_m$ increases, the magnitude of oscillation decreases. 
Additionally, the system's time constant decreases beyond a certain value of $d_m$. 
However, oscillation persists even when $\zeta$ exceeds 1, and the system did not attain 
critical damping. Fig.~\ref{fig:time_responses}(b) shows the results of the same conditions 
as Fig.~\ref{fig:rl2}(b). As the viscosity ratio $d_m$ increases, the oscillation 
becomes smaller, and the time constant becomes lower. Critical damping 
occurs at $\zeta=1$, and no oscillation occurs at higher viscosity ratios. 
The difference between the two conditions is whether (\ref{eq:kam}) is satisfied or not. 
Fig.~\ref{fig:time_responses}(a) has eigenvalues with imaginary components, while 
Fig.~\ref{fig:time_responses}(b) does not. 
The results support the theory that the presence of imaginary components in 
the eigenvalues of the stiffness matrix determines whether the asymmetric matrix 
causes spiral oscillations. 
Since the stiffness matrix in Fig.~\ref{fig:time_responses}(b) only has positive real eigenvalues, 
oscillations can be avoided by setting an appropriate viscosity ratio $d_m$. 
Fig.~\ref{fig:time_responses}(c) compares the paths when $\zeta=1$. 
It shows that the asymmetric matrix with eigenvalues with imaginary components generated 
a spiral path, while the matrix with eigenvalues without imaginary components did not cause 
spiral oscillation.

\subsection{Simulation-based validation of passivity}
\begin{figure}[tb]
    \centering
        \includegraphics[width=8.6cm]{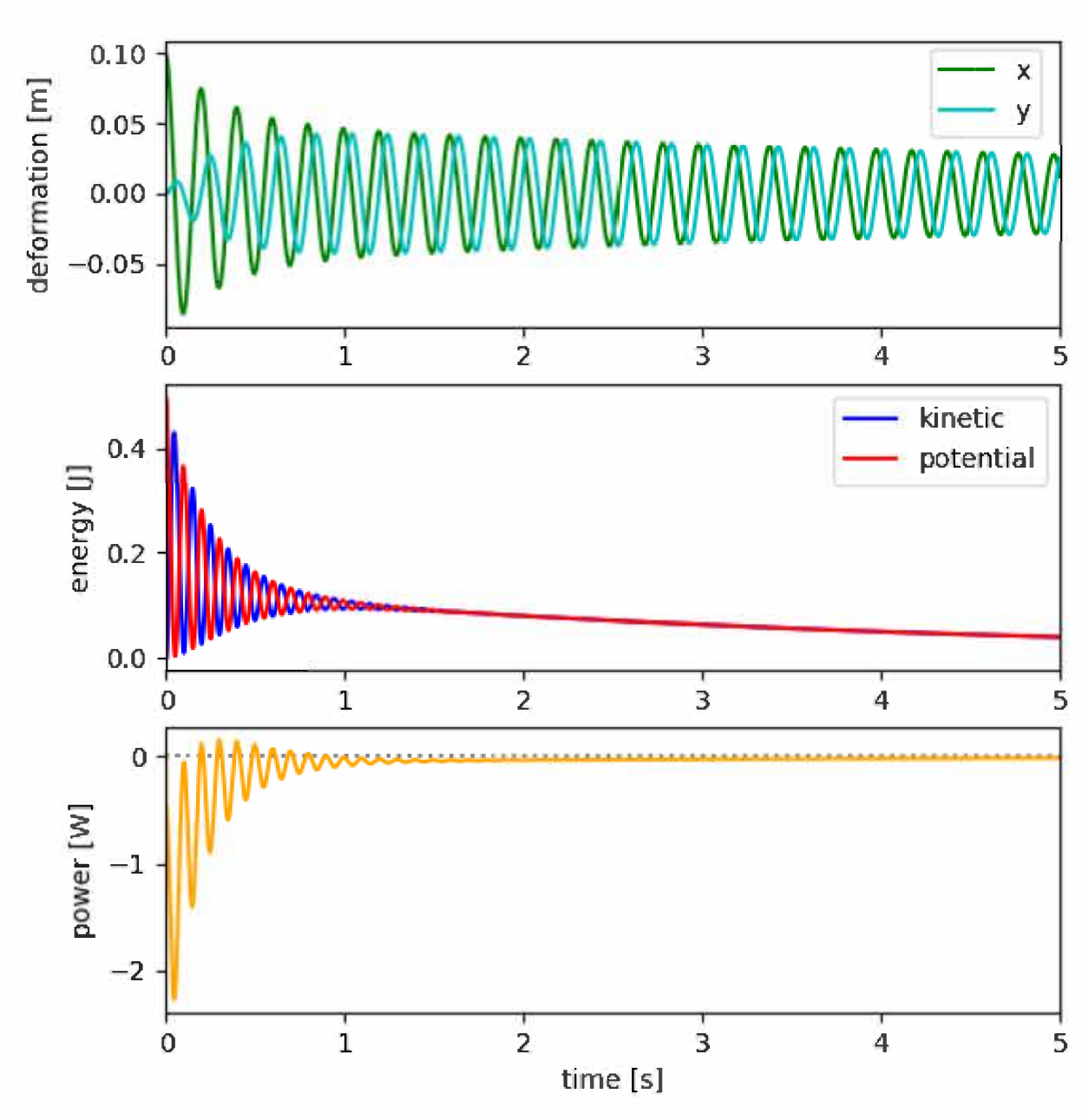}
        \caption{Time response when damping is higher than the threshold (d=0.34)}
        \label{fig:higher_critical}
\end{figure}

\begin{figure}[tb]
    \centering
        \includegraphics[width=8.6cm]{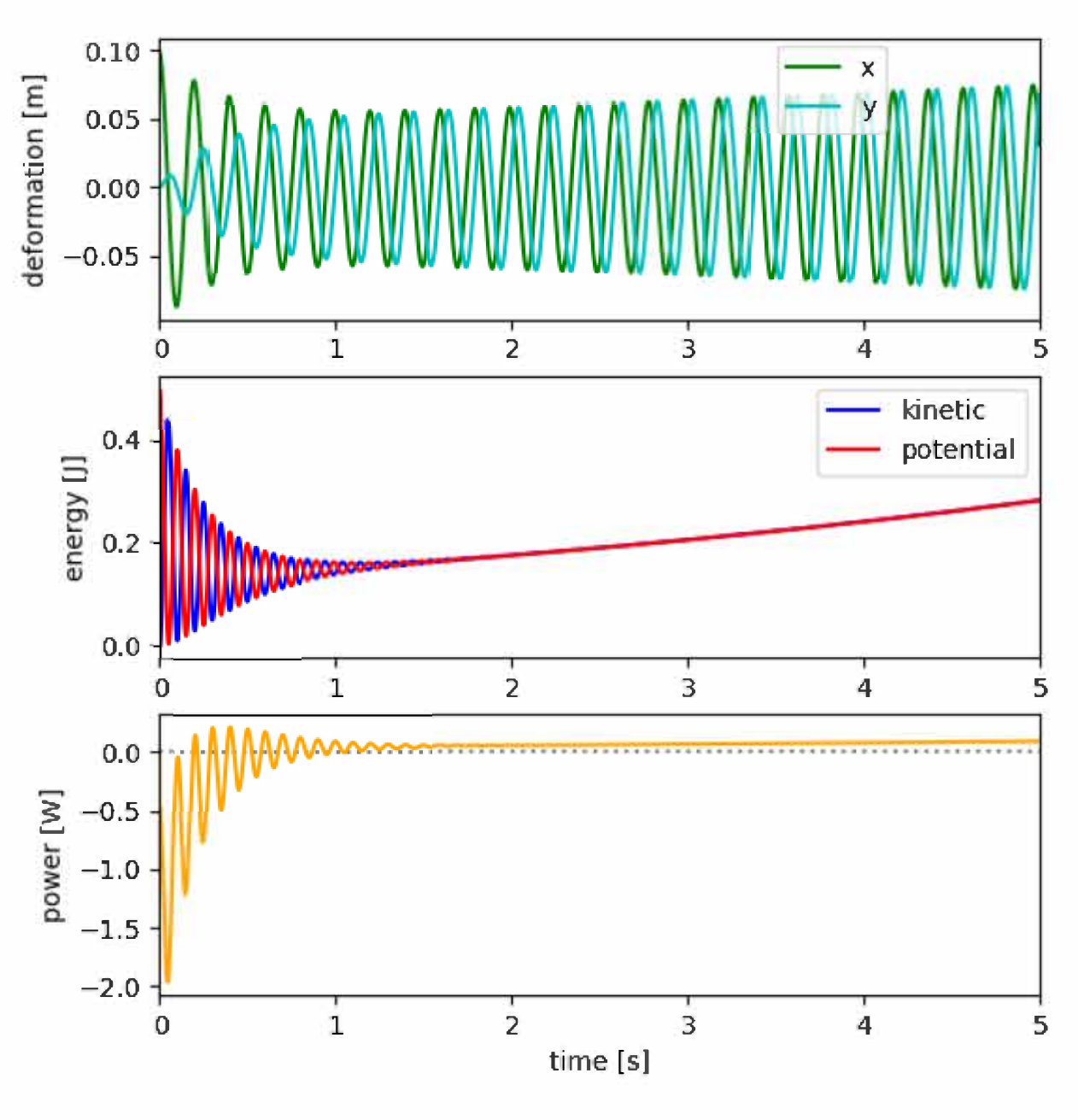}
        \caption{Time response when damping is lower than the threshold (d=0.30)}
        \label{fig:lower_critical}
\end{figure}

The proposed stability analysis method, which utilizes root locus, provides proof of convergence for a particular parameter range. The stability of compliance control is often shown not only by nominal stability, which does not consider contact with the target, but also by combined stability with contact with the target. Therefore, in this simulation, we reproduced a situation in which the object is fixed to a rigid wall.
To illustrate this, we compare the time responses of 
two simulations in Figs.~\ref{fig:higher_critical} and \ref{fig:lower_critical}, with the parameters listed 
in Table~\ref{table:simparam}. 
The parameters of the admittance model are presented in Table~\ref{table:simparam}. 
A damping value of $d=0.34$, which is slightly above the threshold value ($d=0.316$) derived in (\ref{eq:dlimit}), 
caused the system to start oscillating at the mechanical natural frequency and gradually decrease its 
amplitude, as depicted in Fig.~\ref{fig:rl2}. By setting the damping value to $d=0.30$, slightly below the 
threshold value ($d=0.316$), the amplitude of the system gradually increased, 
as shown in Fig.~\ref{fig:lower_critical}. 
This result demonstrates the validity of (\ref{eq:dlimit}). 
Fig.~\ref{fig:higher_critical} shows stable convergence, and the sum of kinetic energy proportional 
to the mass and potential energy proportional to the stiffness of the admittance model gradually decreased. 
However, between 0 and 1 second, there were several time periods when the sum of the outputs was slightly above 0. 
The increase in energy occurred during periods of low velocity during oscillation. 
This means that if the loss due to viscous resistance is smaller than the increase in energy due to the curl field, 
the energy may increase locally even when the amplitude is monotonically decreasing. 
The system does not satisfy local passivity when the velocity is small. 
However, the energy only increases in certain phases during oscillation, and final convergence is guaranteed 
if (\ref{eq:dlimit}) is satisfied.

\begin{table}[tb]
	\centering
	\caption{Parameters in simulation}
	\label{table:simparam}
	\begin{tabular}{l|c|r|l||r|c|r|l}
		\hline
		mass & $m$ & 0.1 & [kg]       &stiffness & $k_x$ & 100.0 & [N/m] \\
		sampling & $t_s$ & 1 & [ms]  && $k_y$ & 100.0 & [N/m] \\
		time & & &                         && $k_s$ & 0.0 & [N/m] \\
		 & & &                               && $k_a$ & 10.0 & [N/m] \\
		\hline
	\end{tabular}
\end{table}


To confirm the validity of the design value of the minimum damper derived in (\ref{eq:dmlimit}), a simulation evaluation was performed: 5 seconds tests were conducted using the same parameters as in Table~\ref{table:simparam}. 
We repeated the test with $k_s$ and $k_a$ changed in the range of -80 to 80. 
The response was recorded when $d$ was 
varied in increments of 0.1 for each condition, and stability was considered to be maintained when the average kinetic energy for 0.1 seconds was less than 0.2 J at the end of the test.
The minimum value of $d$ at which this was the case was plotted. 

\begin{figure}[tb]
    \centering
        \includegraphics[width=8.6cm]{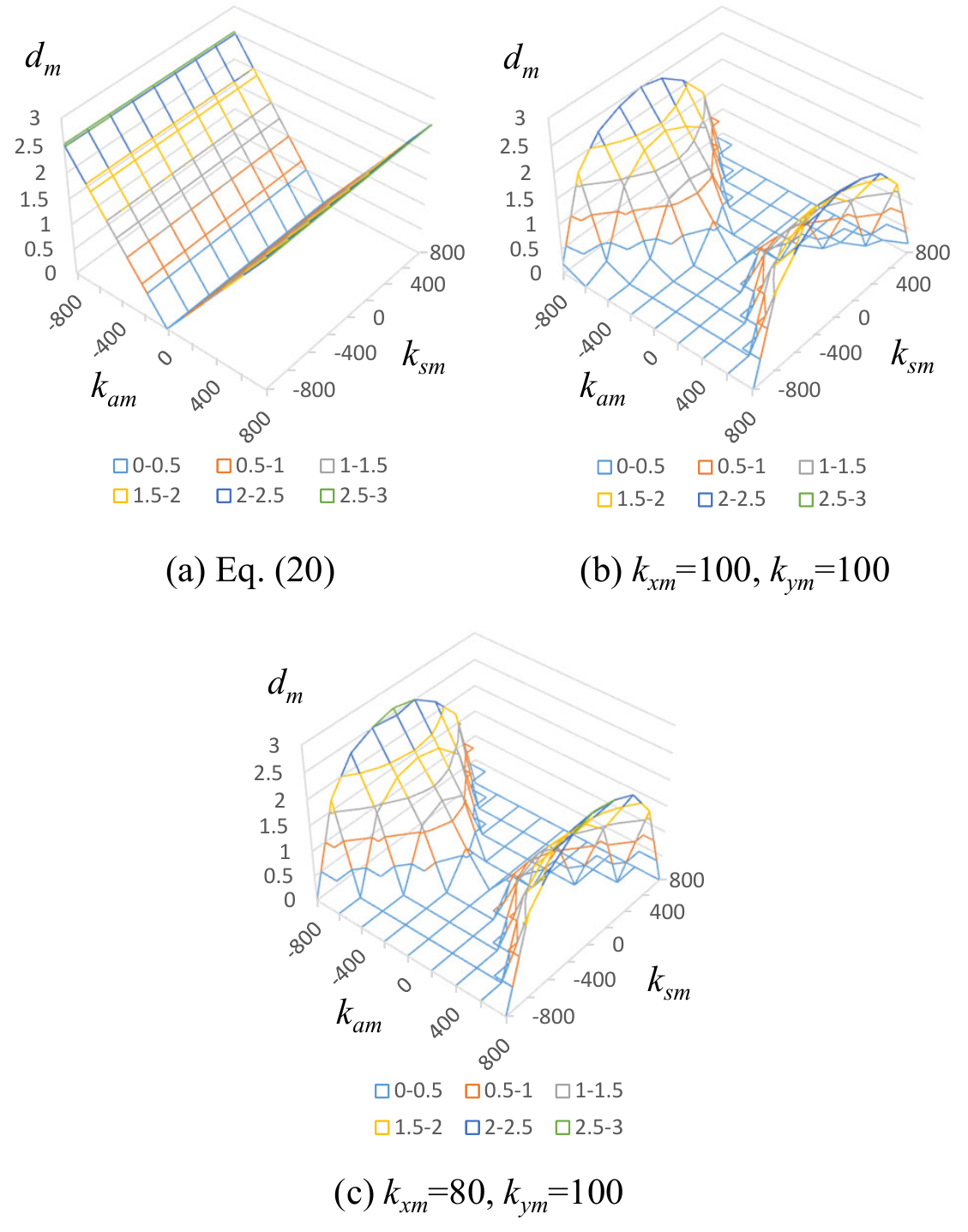}
        \caption{Minimum damper parameter in simulation }
        \label{fig:sim_dm}
\end{figure}

Fig.~\ref{fig:sim_dm}(a) shows the design value derived from (\ref{eq:dmlimit}), and (b) shows the result derived 
from the above simulation. In all conditions, the control system was confirmed to be stable at a damper below the design value. When $k_s$ is 0, the design value and the simulation results almost coincide, while the larger the absolute value of $k_s$, the smaller the damper $d$ becomes. Eq.~(\ref{eq:11}) shows that the imaginary component of the eigenvalue $\lambda$ decreases with increasing absolute value of $k_s$. 
Since the imaginary component of $\lambda$ is the cause of spiral oscillation, this simulation result 
confirms the contents of  (\ref{eq:11}). In addition, the region where stability is ensured even when 
$d$ is zero and the region represented by (\ref{eq:kam}) are almost identical. This confirms the 
validity of (\ref{eq:kam}) as an expression for the parameter region where spiral oscillation does 
not occur. Fig.~\ref{fig:sim_dm}(c) represents the result when only $k_x$ is changed to 80. 
As shown in (\ref{eq:24}), this result confirms that when either stiffness is low, 
it is desirable to set the parameters according to the higher stiffness value. 

The conditions in (\ref{eq:kam}) and (\ref{eq:24}) only concern the parameter settings for the asymmetric part. 
Note that other conditions that must be satisfied by the symmetrical part, as indicated in previous studies, must be taken into account. For example, Fig.~\ref{fig:sim_dm}(b) does not show the results for $k_{sm}$ of 1000, but when $k_{sm}$ 
exceeds 1000, one of the eigenvalues of the symmetric part becomes negative and stability is not satisfied. 
In sum, it is necessary to design the asymmetric part in addition to the design of the symmetric part 
in the conventional methods.

\section{Experiment}
\begin{figure}[tb]
    \centering
        \includegraphics[width=8.6cm]{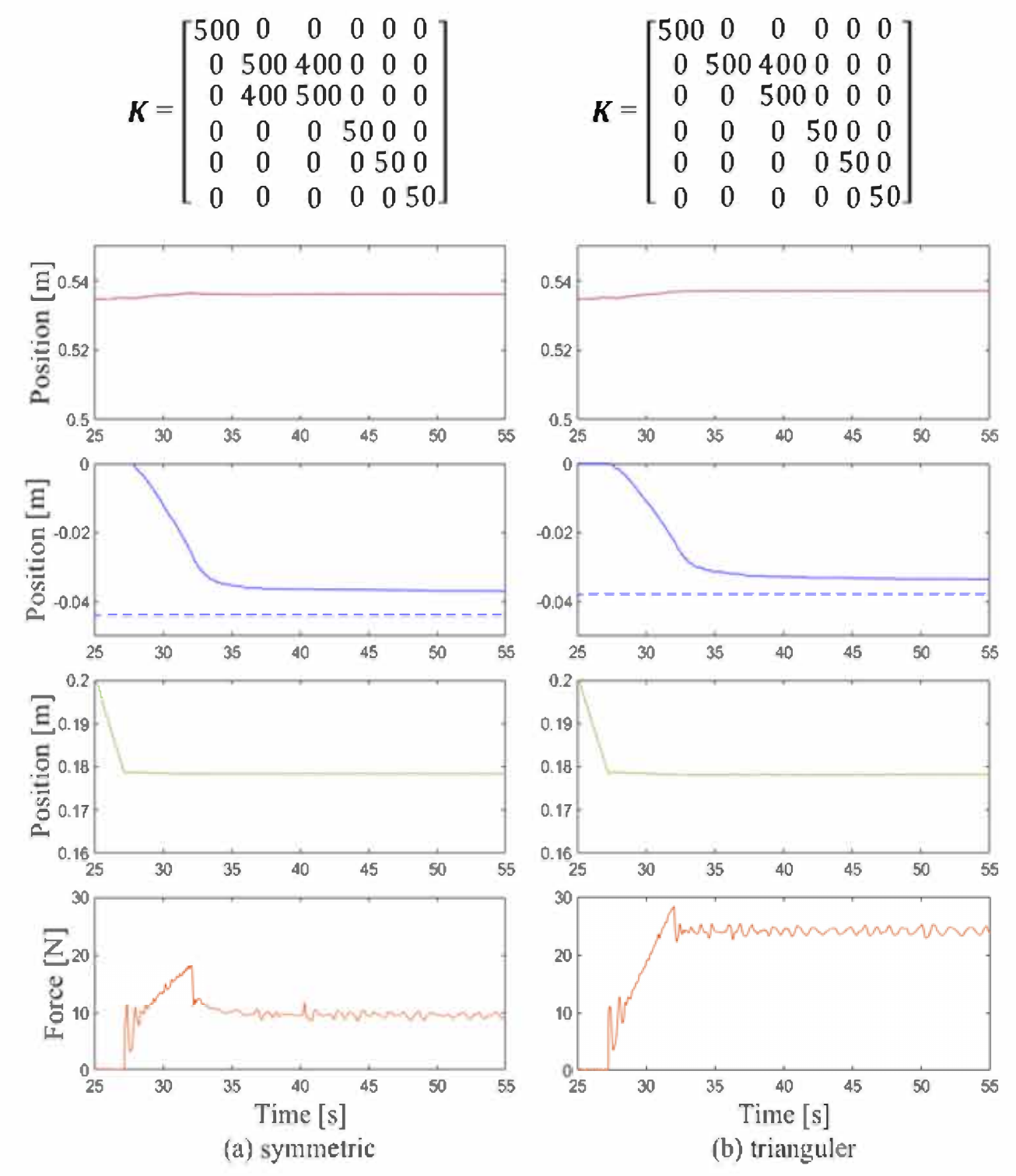}
        \caption{Comparison of experimental results}
        \label{fig:exp}
\end{figure}

In this experiment, the robot's performance was evaluated based on the admittance parameter. 
The robot was made to move vertically towards a metallic board and follow the surface in the y-axis using the displacement caused by the admittance control. Symmetric and asymmetric stiffness matrices were chosen, and damping parameters were set to $d=18$ which was higher than the threshold ($d=17.889$). 
The results shown in Fig.~\ref{fig:exp} revealed that after contacting the board at $t=27$~seconds, the robot moved towards the 
negative direction in the y-axis. There was no noticeable movement in the x-axis due to the design of the non-diagonal element of the stiffness matrix. In the case of the symmetric matrix, the difference between the desired displacement $y_d$ and the actual displacement $y$ was 8 mm. 
Due to the setting of $k_{zy}=400$, there was slight displacement in the z-axis, which led to a lower contact force $f_z$. However, in the case of the asymmetric (triangular) matrix, the difference between the desired displacement $y_d$ and the actual displacement $y$ was 6 mm, which is shorter than that of the symmetric matrix. 
Additionally, the contact force $f_z$ was maintained higher because the displacement mainly occurred in the y-axis.

\section{Conclusion}
This paper discusses the stability of admittance control of a robot arm handling an asymmetric stiffness matrix. 
Although the introduction of asymmetric elements does not guarantee passivity, it is possible to derive a range 
of parameters for which stability is guaranteed from the eigenvalues of the system matrix. 
In addition, the introduction of asymmetric elements may excite helical oscillations, so that critical damping 
may not be obtained for any given damping parameter, depending on the setting of the stiffness matrix. 
In this study, we focused on the fact that the spiral oscillation is caused by the imaginary component of 
the system matrix and clarified the condition of the asymmetric matrix for which critical damping can be obtained. 
From the simulation and experimental results, it is confirmed that the asymmetric matrix setting method 
is effective to obtain critical damping. The analysis was performed on a 2-dimensional plane, but future work 
should extend the theory to a wider range of conditions. The theory can be generalized to systems with three or 
more degrees of freedom and by utilizing the concept of energy tanks, conduct stability analysis similar to a 
passive system. The theory may also be extended to variable stiffness control.

\end{document}